\documentclass[runningheads]{llncs}

\usepackage{placeins}

\usepackage{multirow}
\usepackage{wrapfig}
\renewcommand{\arraystretch}{0.85}

\usepackage[accsupp]{axessibility}  
\usepackage{bm}

\usepackage{graphicx}
\usepackage{booktabs}
\usepackage[pagebackref,breaklinks,colorlinks]{hyperref}

\usepackage{orcidlink}

\begin{document}

\title{SpectraFlow: Unifying Structural Pretraining and Frequency Adaptation for Medical Image Segmentation} 


\author{
Zhiquan Chen \and
Haitao Wang \and
Guowei Zou \and
Hejun Wu
}

\institute{
School of Computer Science and Engineering,
Sun Yat-sen University, Guangzhou, China
\and
Guangdong Key Laboratory of Big Data Analysis and Processing,
Guangzhou, China
\\
\email{wuhejun@mail.sysu.edu.cn}
}

\maketitle

\begin{abstract}
Medical image segmentation remains challenging in low-data regimes, where scarce annotations often yield poor generalization and ambiguous boundaries with missing fine structures.
Recent self-supervised pretraining has improved transferability, but it often exhibits a texture bias. In contrast, accurate segmentation is inherently geometry-aware and depends on both topological consistency and precise boundary preservation.
To address this problem, we propose a two-stage framework that couples structure-aware encoder pretraining with boundary-oriented decoding.
In Stage-1, we aim to learn structure-aware representations for downstream segmentation in low-data regimes. To this end, we propose Mixed-Domain MeanFlow Pretraining, which aligns images and binary masks in a shared latent space through latent transport regression, where masks act as conditional structural guidance rather than prediction targets, making the pretraining task-agnostic. To further improve training stability under scarce supervision, we incorporate a lightweight Dispersive Loss to prevent representation collapse.
In Stage-2, we fine-tune the pretrained encoder with a lightweight decoder that combines Direct Attentional Fusion for adaptive cross-scale gating and Frequency-Directional Dynamic Con\-volution for high-frequency boundary refinement under appearance variation.
Experiments on ISIC-2016, Kvasir-SEG, and GlaS demonstrate consistent gains over state-of-the-art methods, with improved robustness in low-data settings and sharper boundary delineation.
\textbf{The code is in the appendix materials.} 
\keywords{Medical Image Segmentation \and Structure-Aware Pretraining \and Latent Transport \and Boundary Refinement}
\end{abstract}

\begin{figure}[t]
  \centering
  \includegraphics[width=\textwidth]{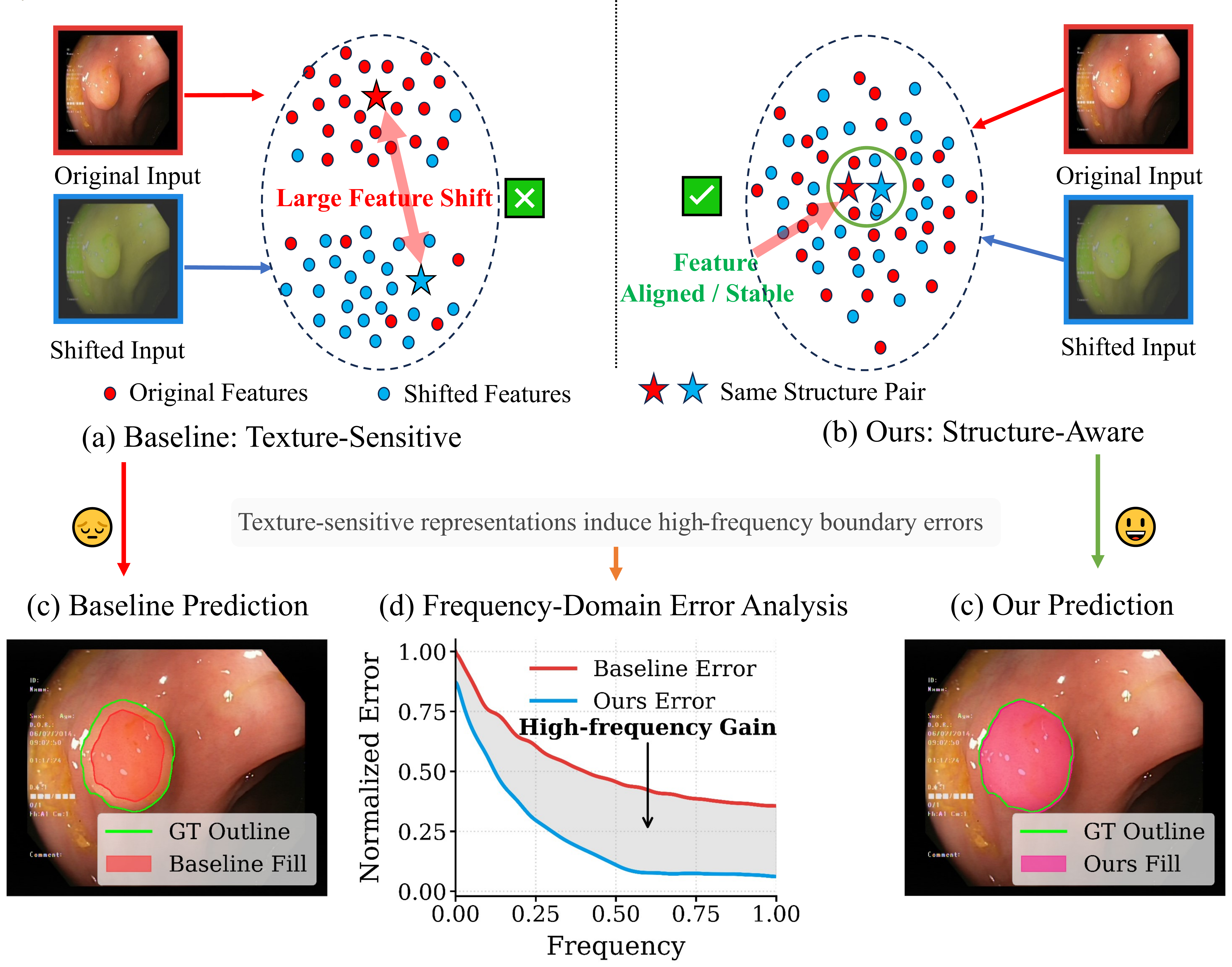}
  \caption{
  \textbf{Motivation: Texture-sensitive representations cause high-frequency boundary errors.}
  \textbf{(a)} Baseline features shift under appearance changes.
  \textbf{(b)} Our features remain domain-invariant.
  \textbf{(c)} Baseline shows boundary misalignment.
  \textbf{(d)} Errors concentrate in high-frequency bands.
  \textbf{(e)} Our method preserves fine boundaries.
}
  \label{fig:motivation}
\end{figure}

\section{Introduction}
Accurate medical image segmentation is essential for computer-aided diagnosis and clinical decision support~\cite{isensee2021nnunet,hatamizadeh2022unetr,chen2021transunet}. 
Such accuracy critically depends on fine anatomical structures. 
However, under low-data supervision, segmentation remains highly challenging. 
Models often suffer from poor generalization, which leads to blurry or broken boundaries, causing fine anatomical structures to be easily lost, as shown in Fig.~\ref{fig:motivation}~\cite{ronneberger2015unet,oktay2018attentionunet,cao2021swinunet}.

A major reason behind this limitation is the texture bias of encoder representations. 
Convolutional encoders, such as U-Net, rely heavily on local appearance patterns. 
These patterns vary significantly across scanners, imaging protocols, and patients, and therefore cannot reliably describe anatomical structures. 
Transformer-based encoders reduce locality bias, but under limited data they may still depend on appearance cues that are easier to learn than geometric structure. 
As a result, models built on both types of encoders are more prone to errors near object boundaries. 
These errors mainly concentrate in mid-to-high frequency components, which encode fine structural details. 
This phenomenon is clearly illustrated in Fig.~\ref{fig:motivation}(d).

Self-supervised pretraining on unlabeled images is widely used to improve representation quality under limited annotations~\cite{he2022mae,bao2022beit,oquab2023dinov2,zhou2021modelsgenesis}. It helps models learn more general visual features. However, most pretraining objectives still focus on reconstructing appearance or predicting tokens. They do not explicitly enforce geometric consistency. In addition, these methods usually depend on large pretraining datasets. In low-data medical settings, both labeled masks and unlabeled images are limited. Therefore, standard self-supervised pretraining cannot fully solve the problem. Segmentation depends strongly on geometry. It requires correct topology, continuous shapes, and precise boundaries. This creates a mismatch between appearance-focused pretraining and geometry-dependent segmentation.

To address this problem, we propose a two-stage framework that shifts learning from texture to geometry and reduces high-frequency boundary errors.  
In Stage-1, we introduce Mixed-Domain Structural Pretraining based on MeanFlow. We embed training images and their binary masks into a shared latent space. We then learn geometry-consistent features through latent-space transport regression. This encourages a smooth transition in representation space so that features change with structure rather than appearance. Masks are used only as conditional structural guidance. They are not prediction targets, and no segmentation loss is used in Stage-1. This keeps pretraining task-agnostic while injecting geometric information. Compared with reconstruction-based pretraining, latent transport regularization better preserves geometric and topological structure. It also does not increase inference cost. To avoid representation collapse and unstable training, we add a lightweight Dispersive Loss to keep features diverse~\cite{wang2025dispersive,caron2021dino,oquab2023dinov2}.

Stage-1 improves encoder representations, but accurate segmentation also requires good boundary recovery in the decoder. Therefore, in Stage-2 we fine-tune the pretrained encoder with a lightweight single-pass decoder that focuses on boundary-related mid-to-high frequency details. A Direct Attentional Fusion module filters noisy skip features and reduces the semantic gap between encoder and decoder. Frequency-Directional Dynamic Convolution then refines boundaries with orientation-aware filtering~\cite{woo2018cbam,chen2020dynamicconv,chen2025frequency}. Stage-1 improves structure awareness in features, and Stage-2 restores fine boundary details during decoding.

The main contributions are:
\begin{itemize}
    \item We show that texture-biased representations are a key reason for poor segmentation in low-data settings and that they lead to high-frequency boundary errors, as shown in Fig.~\ref{fig:motivation}(d).
    \item We propose Mixed-Domain Structural Pretraining with MeanFlow. Masks are used as conditional structural inputs rather than prediction targets. This helps learn geometry-consistent and texture-robust features. A lightweight Dispersive Loss further stabilizes training.
    \item We design an efficient decoder with Direct Attentional Fusion for noise-suppressed feature fusion and Frequency-Directional Dynamic Convolution for fine boundary refinement.
\end{itemize}

\begin{figure}[t!] 
    \centering
    \includegraphics[
        width=\linewidth,
        trim=0cm 0cm 0cm 0cm,
        clip
    ]{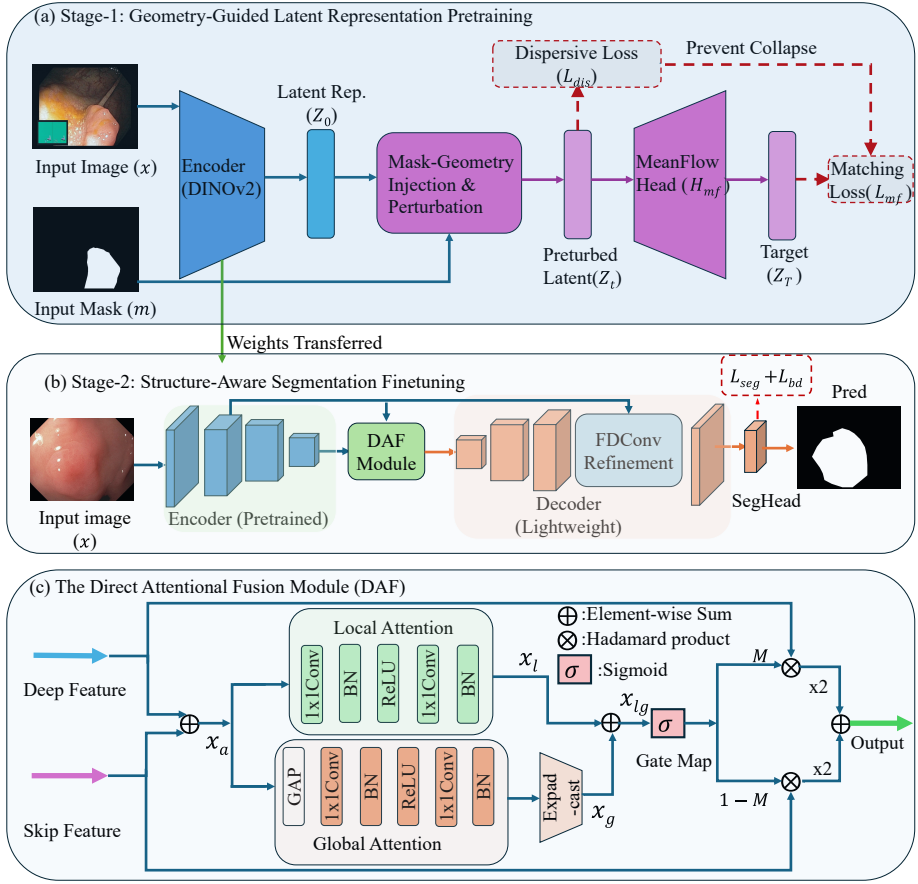}
    \caption{Overview of the proposed two-stage framework.
    (a) Stage-1 performs geometry-guided latent representation pretraining with MeanFlow and Dispersive Loss;
    (b) Stage-2 adopts a lightweight decoder with DAF-based fusion and FDConv refinement;
    (c) illustrates the structure of the DAF module.
    }
    \label{fig:architecture}
\end{figure}

\section{Related Work}
\label{sec:related}

\textbf{Medical image segmentation.}
Encoder--decoder architectures remain the dominant paradigm for medical image segmentation, covering both CNN- and trans\-former-based designs~\cite{ronneberger2015unet,isensee2021nnunet,zhou2019unetplusplus,hatamizadeh2022unetr}. 
Although recent methods improve contextual aggregation and feature fusion, many still require substantial labeled data for reliable training~\cite{ronneberger2015unet,isensee2021nnunet}. 
Under limited annotations, models tend to overfit to appearance and texture cues~\cite{Geirhos2019TextureBias,Hermann2019TextureOrigins}, which hinders generalization across scanners and patients~\cite{Guan2022DomainShiftSurvey,Ugurlu2021DomainShiftSeg}, especially when boundary contrast is weak. 
This limitation motivates representation learning with stronger structural priors~\cite{zhou2019unetplusplus}.

\textbf{Representation pretraining for dense prediction.}
Self-supervised pretraining, including masked image modeling (MIM) and contrastive learning, has shown strong transferability in vision~\cite{he2022mae,chen2020simclr,he2020moco,caron2021dino,bao2022beit}. 
However, most MIM objectives rely on pixel-level reconstruction and thus emphasize local texture recovery~\cite{Geirhos2019TextureBias,Hermann2019TextureOrigins}, without explicitly modeling the geometric or structural manifold required for dense segmentation~\cite{he2022mae,bao2022beit}. 
Transport-based learning instead organizes representations through trajectories in latent space, but flow- and optimal-transport-style pretraining remain underexplored in medical image segmentation~\cite{zhao2023lfd}. 
Our Stage-1 addresses this gap by using latent optimal transport to learn geometry-consistent representations, which is particularly helpful in low-data regimes.

\textbf{Boundary-aware and frequency-based modeling.}
To enhance boundary quality, prior works incorporate frequency cues such as Fourier or wavelet transforms, or use attention-guided decoder fusion~\cite{lee2022fnet,fujieda2018waveletcnn,woo2018cbam}. 
However, many frequency-domain designs are global or isotropic and rely on fixed operators, which limits their sensitivity to local boundary orientation and contextual variation~\cite{lee2022fnet,fujieda2018waveletcnn}. 
Our Stage-2 introduces a lightweight frequency-directional dynamic convolution that adaptively aggregates context with orientation awareness, together with a gating fusion module to suppress noisy skip features. This leads to sharper boundary delineation and fewer false positives~\cite{chen2020dynamicconv,woo2018cbam}.

\section{Network Architecture}
\label{sec:architecture}

An overview of the proposed two-stage framework is shown in Fig.~\ref{fig:architecture}.
Stage-1 conducts MeanFlow-based representation pretraining in the encoder latent space with explicit mask-geometry injection and Dispersive Loss regularization, while Stage-2 employs a lightweight decoder with DAF-based gated skip fusion and FDConv refinement for supervised segmentation.
This section outlines the overall architecture, and detailed formulations are provided in Sec.~\ref{sec:method}.

\textbf{Stage-1: Encoder Pretraining.}
Given an input image $\mathbf{x}$, the encoder produces a latent representation $\mathbf{z}_0 = E(\mathbf{x})$.
Masks are used only as structure-only inputs for latent shaping, and no segmentation loss is applied in Stage-1.
MeanFlow pretraining is performed on perturbed latents $\mathbf{z}_t$, where mask geometry and a \emph{Dispersive Loss} are used to regularize the latent space and promote informative representations (Sec.~\ref{sec:meanflow}).
The MeanFlow head is discarded after Stage-1.

\textbf{Stage-2: Decoder Finetuning.}
For supervised segmentation, a lightweight decoder progressively upsamples encoder features.
DAF performs gated fusion on a mid-level skip connection, followed by FDConv-based frequency--directional refinement to enhance boundary details (Sec.~\ref{sec:fdconv}).

\section{Method}
\label{sec:method}

\subsection{Stage-1: MeanFlow Pretraining}
\label{sec:meanflow}

\textbf{Motivation and Usage.}
MeanFlow~\cite{geng2025meanflow} is an efficient framework for latent-space optimal transport, originally realized with representation autoencoders~\cite{hu2025meanflow_rae}.
In this work, we repurpose MeanFlow \emph{solely} as an encoder pretraining objective for medical image segmentation, aiming to shape a structured latent representation in Stage-1 without introducing any additional inference cost.
Crucially, Stage-1 focuses on \emph{structure-aware representation learning}: masks, when used, serve exclusively as conditional structural inputs (\emph{not} prediction targets).
We optimize the encoder via latent-space transport regression rather than pixel-level reconstruction, promoting geometric and topological consistency essential for dense segmentation tasks. Accordingly, Stage-1 aims to learn \emph{class-agnostic} structural priors (shape continuity, topology, and boundary layout) rather than class-specific semantics.

\textbf{Latent Representation and Perturbation.}
Given an input sample $\mathbf{x}\in\mathbb{R}^{3\times H\times W}$, the encoder $E(\cdot)$ produces a latent feature map
$\mathbf{z}_0 = E(\mathbf{x}) \in \mathbb{R}^{C\times h\times w}$.
MeanFlow pretraining is performed in the latent space by constructing perturbed latents:
\begin{equation}
\mathbf{z}_t = \alpha_t \mathbf{z}_0 + \sigma_t \bm{\epsilon}, 
\quad \boldsymbol{\epsilon}\sim\mathcal{N}(\mathbf{0},\mathbf{I}), \quad t\in[0,1],
\label{eq:latent_perturb}
\end{equation}
where we employ a linear schedule $\alpha_t = 1-t$ and $\sigma_t = t$.
This perturbation process encourages representations to remain stable and consistent under controlled latent noise, a desirable property for segmentation in low-data regimes.

\textbf{MeanFlow Objective for Encoder Pretraining.}
Instead of predicting an instantaneous velocity, MeanFlow models a long-step transport direction, defined as the \emph{time-averaged} velocity between two time points $t>s$:
\begin{equation}
h_\theta(\mathbf{z}_t,t,s)
\approx 
\frac{1}{t-s}\int_{s}^{t} v(\mathbf{z}_u,u)\,du,
\label{eq:meanflow_def}
\end{equation}
where $v(\cdot)$ denotes the underlying latent dynamics induced by the perturbation.
Following the self-distilled MeanFlow identity~\cite{geng2025meanflow}, we train the estimator $h_\theta$ using a stop-gradient regression objective:
\begin{equation}
\mathcal{L}_{\text{MF}}
=
\mathbb{E}_{t>s}\mathbb{E}_{\mathbf{z}_t}
\Big[
\big\|
h_\theta(\mathbf{z}_t,t,s)
-
h^{\text{tgt}}_{\theta^-}(\mathbf{z}_t,t,s;\,w)
\big\|_2^2
\Big],
\label{eq:meanflow_loss}
\end{equation}
where $h^{\text{tgt}}_{\theta^-}$ is the target generated by a stop-gradient network and $w$ is a proxy velocity.
Following~\cite{geng2025meanflow}, $w$ is computed from the target network to form a self-distilled regression target.
Notably, Stage-1 is driven purely by latent-space transport regression: any masks involved are utilized strictly as conditional structural inputs, ensuring the encoder learns class-agnostic structural priors rather than overfitting to task-specific semantics.

\textbf{Dispersive Regularization.}
\label{sec:dis}
Although MeanFlow provides strong representation learning signals, latent features of medical images often exhibit high appearance similarity, which can lead to overly compact distributions and reduced feature separability. This potential feature collapse may destabilize the transport regression.
To address this, we introduce a lightweight \emph{Dispersive Loss} that treats all in-batch representations as negative pairs, explicitly enlarging inter-sample distances in the latent space.
This dispersion mechanism requires neither positive pairs nor additional annotations, introducing negligible computational overhead.

The overall Stage-1 objective is formulated as:
\begin{equation}
\mathcal{L}=\mathcal{L}_{\text{MF}}+\lambda\,\mathcal{L}_{\text{Disp}}.
\label{eq:mf_disp}
\end{equation}
Adopting the ``contrastive-style objective without explicit positives'' paradigm~\cite{wang2025dispersive}, we instantiate $\mathcal{L}_{\text{Disp}}$ using a repulsive term based on squared $\ell_2$ distances.
Let $\{h_i\}_{i=1}^{B}$ denote the global average pooled features from the penultimate encoder block within a mini-batch, and let $\tau>0$ be a temperature parameter. We define:
\begin{equation}
\mathcal{L}_{\mathrm{Disp}}
= 
\log\Bigg(
\frac{1}{B(B-1)}\sum_{i\neq j}
\exp\Big(-\|h_i-h_j\|_2^2/\tau\Big)
+\varepsilon
\Bigg),
\label{eq:disp_infonce_l2}
\end{equation}
where $h_i \in \mathbb{R}^d$ is the representation of the $i$-th sample.
Minimizing Eq.~\eqref{eq:disp_infonce_l2} reduces in-batch feature similarity and encourages a more dispersed representation space, mitigating collapse.
$\varepsilon$ is a small constant for numerical stability.

\textbf{Mixed-Domain Latent Pretraining.}
\label{sec:mixed_domain}
Stage-1 adopts a Mixed-Domain Latent Pretraining strategy, where raw medical images and binary masks from the training set are optimized jointly.
Binary masks serve as structure-only inputs that emphasize anatomical layout and boundary geometry, complementing the appearance-rich semantics of raw images.
To align dimensions, we repeat the single-channel mask along the channel dimension to match the encoder input format.
Importantly, masks are used strictly as additional inputs, \emph{never} as prediction targets or supervision signals.
Image and mask inputs are mixed with a fixed sampling ratio to form Stage-1 mini-batches.
The impact of this strategy is analyzed in our ablation study.

\subsection{Stage-2: Segmentation Finetuning}
\label{sec:stage2_seg}

After MeanFlow pretraining, we attach a lightweight segmentation decoder $D(\cdot)$ and finetune the model on labeled data:
\begin{equation}
\hat{\mathbf{y}} = D(E(\mathbf{x})).
\end{equation}
The decoder is initialized from scratch, while the encoder inherits weights from Stage-1. To balance representation adaptation and training stability, we unfreeze only the last encoder block (keeping earlier blocks frozen).
The model is optimized via a composite loss comprising a region-based term and a boundary-aware term:
\begin{equation}
\mathcal{L}_{\text{S2}}
=
\mathcal{L}_{\text{seg}}(\hat{\mathbf{y}},\mathbf{y})
+
\beta\,\mathcal{L}_{\text{bd}}(\hat{\mathbf{y}},\mathbf{y}),
\label{eq:stage2_loss}
\end{equation}
where $\mathcal{L}_{\text{seg}}$ combines Dice and BCE losses, and $\mathcal{L}_{\text{bd}}$ penalizes boundary errors to delineate fine anatomical structures.

To further refine boundaries, we design the decoder with two key components in a coarse-to-fine manner:
First, Direct Attentional Fusion (DAF) injects mid-level encoder features via a gated mechanism to bridge the semantic gap.
Second, Frequency-Directional Dynamic Convolution (FDConv) is applied to the fused features to strengthen oriented high-frequency responses.
This ordering (DAF $\rightarrow$ FDConv) ensures FDConv operates on detail-enriched representations while maintaining an efficient single-pass inference flow.

\textbf{Cross-Scale Feature Injection via Direct Attentional Fusion.}
\label{sec:daf}
While the pretrained encoder captures robust high-level semantics, deep features often lack the fine-grained spatial details required for precise boundary delineation.
To recover these details, we introduce a long-range skip connection injecting intermediate encoder features into the decoder.
However, naive concatenation is suboptimal due to the semantic gap: shallow encoder features may contain background noise, while decoder features are semantically consistent but spatially coarse.
We therefore propose the Direct Attentional Fusion (DAF) module to perform content-adaptive gated fusion.

Let $\mathbf{f}_{\text{skip}}$ be the projected encoder feature aligned to the decoder dimension.
We first compute a joint representation $\mathbf{f}_{\text{sum}} = \mathbf{f}_{\text{dec}} + \mathbf{f}_{\text{skip}}$.
DAF dynamically modulates this integration by aggregating local and global contexts via dual pathways: a local branch $\mathcal{B}_{\text{local}}$ (utilizing point-wise convolution) for texture calibration and a global branch $\mathcal{B}_{\text{global}}$ (utilizing channel attention) for semantic alignment.
A dense gating mask $\mathbf{M} \in [0, 1]^{C \times H \times W}$ is generated by combining these branches:
\begin{equation}
\mathbf{M}
=
\sigma \Big(
\mathcal{B}_{\text{local}}(\mathbf{f}_{\text{sum}})
+
\mathcal{B}_{\text{global}}(\mathbf{f}_{\text{sum}})
\Big).
\label{eq:daf_mask}
\end{equation}
This mask acts as a content-adaptive gate, regulating fusion via a complementary formulation:
\begin{equation}
\mathbf{f}_{\text{fused}}
=
2 \cdot \mathbf{f}_{\text{dec}} \odot \mathbf{M}
+
2 \cdot \mathbf{f}_{\text{skip}} \odot (\mathbf{1} - \mathbf{M}).
\label{eq:daf_fusion}
\end{equation}
By suppressing discordant noise from shallow layers while selectively enhancing boundary-aware cues, DAF harmonizes cross-scale representations for subsequent refinement.

\textbf{FDConv: Frequency--Directional Refinement.}
\label{sec:fdconv}
FDConv is adopted as a plug-in refinement operator to enhance boundary-sensitive high-frequency features.
Following~\cite{chen2025frequency}, FDConv constructs a sample-adaptive $3{\times}3$ kernel by modulating learnable frequency-domain parameters with attention signals derived from the input.
Specifically, let $\widetilde{\mathbf{W}}$ denote the learnable complex coefficients in the frequency domain.
Given feature map $\mathbf{f}$, FDConv predicts modulation weights $\mathbf{a}(\mathbf{f})$ and synthesizes an effective kernel via inverse FFT:
\begin{equation}
\mathbf{W}_{\mathrm{eff}}(\mathbf{f})
=
\mathcal{F}^{-1}\!\Big(
\mathbf{a}(\mathbf{f}) \odot \widetilde{\mathbf{W}}
\Big),
\label{eq:fdconv_kernel_compact}
\end{equation}
where $\odot$ denotes element-wise modulation with broadcasting.
The output is produced by standard convolution:
\begin{equation}
\mathrm{FDConv}_{3\times3}(\mathbf{f}) = \mathbf{W}_{\mathrm{eff}}(\mathbf{f}) * \mathbf{f}.
\label{eq:fdconv_conv_compact}
\end{equation}
We refer readers to~\cite{chen2025frequency} for implementation details. In our framework, FDConv replaces the first $3{\times}3$ convolution in the decoder refinement block to explicitly sharpen high-frequency boundary details.

\section{Experiments}
\label{sec:experiments}

\subsection{Datasets and Metrics}
\label{sec:datasets_metrics}

To evaluate the effectiveness and generality of our method for medical image segmentation, we conduct experiments on three widely used benchmarks: Kvasir\mbox{-}SEG~\cite{Jha2020KvasirSEG}, GlaS~\cite{Sirinukunwattana2017GlaS}, and ISIC\mbox{-}2016~\cite{Gutman2016ISIC}.
These datasets span representative medical imaging modalities and segmentation subtasks, including colonoscopic polyp segmentation, histopathology gland segmentation, and dermoscopic skin lesion segmentation, enabling a comprehensive evaluation across diverse appearance patterns and boundary characteristics.

We follow the official evaluation protocols for all datasets.
Specifically, Kvasir\-\mbox{-}SEG uses an 880/120 split for training and validation.
GlaS follows the official 85/80 split for training and validation.
ISIC\mbox{-}2016 adopts the official split of 900/379 for training and validation.
Across all datasets, the foreground corresponds to the target anatomical structure (polyp, gland, or skin lesion), while all remaining pixels are treated as background.

We report standard medical image segmentation metrics, including mIoU, mDSC, Recall, Precision, and HD95.
mIoU and mDSC measure region overlap, while HD95 evaluates boundary accuracy based on the 95th percentile of the symmetric Hausdorff distance.
For all experiments, input images are resized to $224 \times 224$ pixels for both training and evaluation.
Consequently, all metrics are computed per image at this resolution and averaged over the evaluation set; results are reported in percentage unless otherwise stated, with HD95 measured in pixels.

\subsection{Implementation Details}
\label{sec:implementation}

We implement our framework using PyTorch.
All experiments are conducted on a single NVIDIA RTX 4090 GPU.
Stage-1 and Stage-2 are trained sequentially following the two-stage protocol described in Sec.~\ref{sec:method}.

\textbf{Stage-1: MeanFlow Pretraining.}
In Stage-1, the encoder is pretrained using the MeanFlow objective for 300 epochs.
We instantiate the encoder $E(\cdot)$ as a DINOv2-based vision transformer and perform latent-space transport learning on features of spatial resolution $16{\times}16$.
MeanFlow-based transport and dispersive regularization are applied to optimize the encoder representations, without modifying the backbone architecture.
The model is trained using AdamW with an initial learning rate of $1\times10^{-5}$, linearly decayed to $1\times10^{-6}$.
The global batch size is set to 8 with gradient accumulation, and an exponential moving average with decay rate 0.999 is used for training stabilization.

\textbf{Stage-2: Segmentation Finetuning.}
In Stage-2, a lightweight decoder is attached and trained for up to 100 epochs.
The decoder is trained from scratch, while the encoder is initialized from Stage-1 and finetuned by unfreezing only the last encoder block.
Early stopping is applied based on validation Dice score.
The learning rate is adaptively reduced by a factor of 0.5 when validation performance plateaus, with a minimum learning rate of $1\times10^{-6}$.
Unless otherwise specified, consistent training settings are used across all datasets.

\begin{table}[t]
  \centering
  \caption{Comparison with other methods on the ISIC-2016, Kvasir-SEG, and GlaS datasets. 
  \textbf{Bold} indicates best performance, and \underline{underline} indicates second-best.}
  \label{tab:compare1}

  \setlength{\tabcolsep}{1.5pt} 
  \renewcommand{\arraystretch}{1.15}
  
  \resizebox{\textwidth}{!}{%
  \begin{tabular}{l cccc cccc cccc}
    \toprule
    \multirow{2}{*}{Methods}
      & \multicolumn{4}{c}{ISIC-2016}
      & \multicolumn{4}{c}{Kvasir-SEG}
      & \multicolumn{4}{c}{GlaS} \\
    \cmidrule(lr){2-5} \cmidrule(lr){6-9} \cmidrule(lr){10-13}
      & mIoU & mDSC & Rec. & Prec.
      & mIoU & mDSC & Rec. & Prec.
      & mIoU & mDSC & Rec. & Prec. \\
    \midrule
    U\mbox{-}Net (MICCAI 2015) 
      & 83.61 & 90.32 & 88.43 & 89.45 
      & 65.54 & 75.81 & 83.62 & 77.63 
      & 75.83 & 85.51 & 90.34 & 82.82 \\
    U\mbox{-}Net++ (MICCAI 2018) 
      & 83.54 & 89.89 & 89.76 & 90.42
      & 67.93 & 77.24 & 86.57 & 77.78
      & 77.61 & 86.97 & 89.65 & 85.56 \\
    Attn U\mbox{-}Net (MIDL 2018) 
      & 83.41 & 90.23 & 88.56 & 89.77 
      & 67.64 & 77.43 & 83.95 & 79.94 
      & 76.63 & 85.94 & 91.80 & 82.20 \\
    PraNet (MICCAI 2020) 
      & 84.41 & 91.12 & 90.74 & 91.98
      & 83.04 & 89.40 & 90.62 & 91.37 
      & 71.80 & 83.09 & 90.90 & 78.05 \\
    TGANet (MICCAI 2022) 
      & 82.43 & 90.12 & 91.10 & 90.84 
      & 83.30 & 89.24 & 91.31 & 91.25 
      & 77.11 & 81.84 & 84.72 & 80.23 \\
    DCSAU\mbox{-}Net (CBM 2023) 
      & 85.31 & 91.90 & 91.14 & 90.79
      & 83.51 & 88.92 & 89.53 & 89.50 
      & 77.63 & 86.51 & 93.08 & 82.55 \\
    XBFormer (TMI 2023) 
      & 84.46 & 90.82 & 91.04 & 90.88 
      & 83.81 & 89.07 & 89.80 & 87.22 
      & 73.70 & 84.33 & 84.02 & 85.75 \\
    CASF\mbox{-}Net (CBM 2023) 
      & 81.45 & 88.66 & 89.12 & 88.56 
      & 81.71 & 88.72 & 89.24 & 88.24 
      & 78.43 & 87.21 & 91.35 & 85.94 \\
    DTAN (KBS 2024) 
      & 81.14 & 87.56 & 89.41 & 88.23 
      & 84.10 & 90.44 & \underline{91.65} & \underline{92.06} 
      & 78.55 & 87.90 & 88.51 & 90.23 \\
    \mbox{DoubleAANet (IF 2025)}
      & 85.14 & 91.48 & 92.16 & \underline{92.24} 
      & 82.12 & 90.41 & 91.32 & 76.38 
      & 83.28 & 85.60 & 88.14 & 90.16 \\
    ConDSeg (AAAI 2025)
      & \underline{86.28} & \underline{92.24} & \textbf{93.25} & 90.66 
      & \underline{84.62} & \underline{90.45} & \textbf{92.17} & 91.55
      & \underline{84.96} & \underline{91.38} & \underline{93.17} & \underline{90.24} \\
    \midrule
    \textbf{Ours} 
      & \textbf{86.88} & \textbf{92.98} & \underline{93.22} & \textbf{92.74} 
      & \textbf{85.90} & \textbf{91.60} & 91.30 & \textbf{93.88} 
      & \textbf{85.63} & \textbf{92.12} & \textbf{93.28} & \textbf{91.20} \\
    \bottomrule
  \end{tabular}%
  }
\end{table}

\subsection{Comparison with State-of-the-Art Methods}
\label{sec:sota_comparison}

Table~\ref{tab:compare1} reports comparisons with representative CNN- and transformer-based methods on ISIC-2016, Kvasir-SEG, and GlaS. 
Our method consistently achieves the best performance across all three datasets, indicating strong robustness under diverse imaging characteristics.

On ISIC-2016 and Kvasir-SEG, it achieves the highest mIoU/mDSC and the best precision with comparable recall, suggesting fewer false positives without loss of sensitivity.
On GlaS, where annotations are scarce and gland structures are densely distributed, our method yields the best results on all metrics, further validating its generalization in low-data regimes.

\begin{table}[t]
\centering
\small
\caption{
\textbf{Ablation of Stage-1 Pretraining Strategies} on ISIC-2016.
All methods share the same Stage-2 finetuning setting.
\emph{Mixed-domain MAE} replaces MeanFlow with masked reconstruction under identical inputs.
Masks are used only as conditional structural inputs, never as prediction targets.
}

\label{tab:ablation_pretrain}
\setlength{\tabcolsep}{2.0pt} 
\begin{tabular}{llccc}
\toprule
\textbf{Stage-1 Strategy} & \textbf{Pretrain Data} & \textbf{Dice (\%)} & \textbf{mIoU (\%)} & \textbf{HD95 $\downarrow$} \\
\midrule
No Stage-1 (Official DINOv2)
& Image
& 80.12 & 70.14 & 34.18 \\
MeanFlow (Image-only) 
& Image 
& 82.15 & 71.20 & 21.45 \\

Mixed-domain MAE
& Image + Mask 
& 85.50 & 75.25 & 19.86 \\

+ Mixed-domain MeanFlow 
& Image + Mask 
& 87.10 & 78.40 & 17.80 \\

\textbf{+ Dis (Ours)} 
& Image + Mask 
& \textbf{88.62} & \textbf{80.55} & \textbf{17.24} \\

\bottomrule
\end{tabular}
\end{table}

\begin{table}[t]
\centering
\small
\caption{
\textbf{Ablation of Fusion Strategies and Attention Modules (Stage-2).}
``Standard'' denotes the baseline residual block with $3{\times}3$ convolutions.
FDConv replaces the first $3{\times}3$ convolution (\texttt{conv1}) in the refinement block, while CBAM is inserted \emph{within} the block before residual addition.
HD95 is reported in pixels at $224 \times 224$ resolution.
}
\label{tab:ablation_cbam}
\setlength{\tabcolsep}{2.0pt}
\begin{tabular}{lccccc}
\toprule
\textbf{Method} & \textbf{Fusion} & \textbf{Decoder Block} & \textbf{Dice (\%)} & \textbf{mIoU (\%)} & \textbf{HD95 $\downarrow$} \\
\midrule
Baseline & Concat & Standard & 88.62 & 80.55 & 17.24 \\
+ CBAM & Concat & Standard + CBAM & 87.80 & 80.12 & 17.56 \\
\textbf{+ FDConv (Ours)} & Concat & \textbf{FDConv} & 91.65 & 85.92 & 12.15 \\
\textbf{+ DAF (Ours)} & \textbf{DAF} & Standard & 91.50 & 85.64 & 12.34 \\
+ DAF + CBAM & DAF & Standard + CBAM & 88.36 & 79.92 & 15.54 \\
\midrule
\textbf{Full (Ours)} & \textbf{DAF} & \textbf{FDConv} & \textbf{92.98} & \textbf{86.88} & \textbf{10.86} \\
\bottomrule
\end{tabular}
\end{table}

\subsection{Ablation Studies}
\label{sec:ablation}

We conduct ablation studies to quantify the contribution of each component under the same evaluation protocol.

\textbf{Stage-1 pretraining strategies.}
Table~\ref{tab:ablation_pretrain} first establishes a strong baseline by directly fine-tuning the official DINOv2 encoder without Stage-1 pretraining, which yields limited performance (Dice $80.12\%$), indicating that generic self-supervised features alone are insufficient for accurate medical image segmentation.
Replacing transport-based pretraining with a mixed-domain masked autoencoding objective improves performance (Dice $85.50\%$), suggesting that incorporating structural inputs is beneficial but still limited by reconstruction-centric supervision.
Building upon this baseline, a clear progression is observed from image-only MeanFlow pretraining (Dice $82.15\%$) to mixed-domain MeanFlow with structure-only masks (Dice $87.10\%$), and further to Dispersive Regularization (Dice $88.62\%$), demonstrating that structural inputs effectively complement appearance features and that dispersion stabilizes transport-based representation learning.
Unless otherwise specified, the Stage-1 model in the last row of Table~\ref{tab:ablation_pretrain} is used to initialize Stage-2; hence the \emph{Baseline} in Table~\ref{tab:ablation_cbam} corresponds to this pretrained encoder.

\begin{wrapfigure}{r}{0.45\textwidth}
    \centering
    \vspace{-25pt} 
    \includegraphics[width=\linewidth]{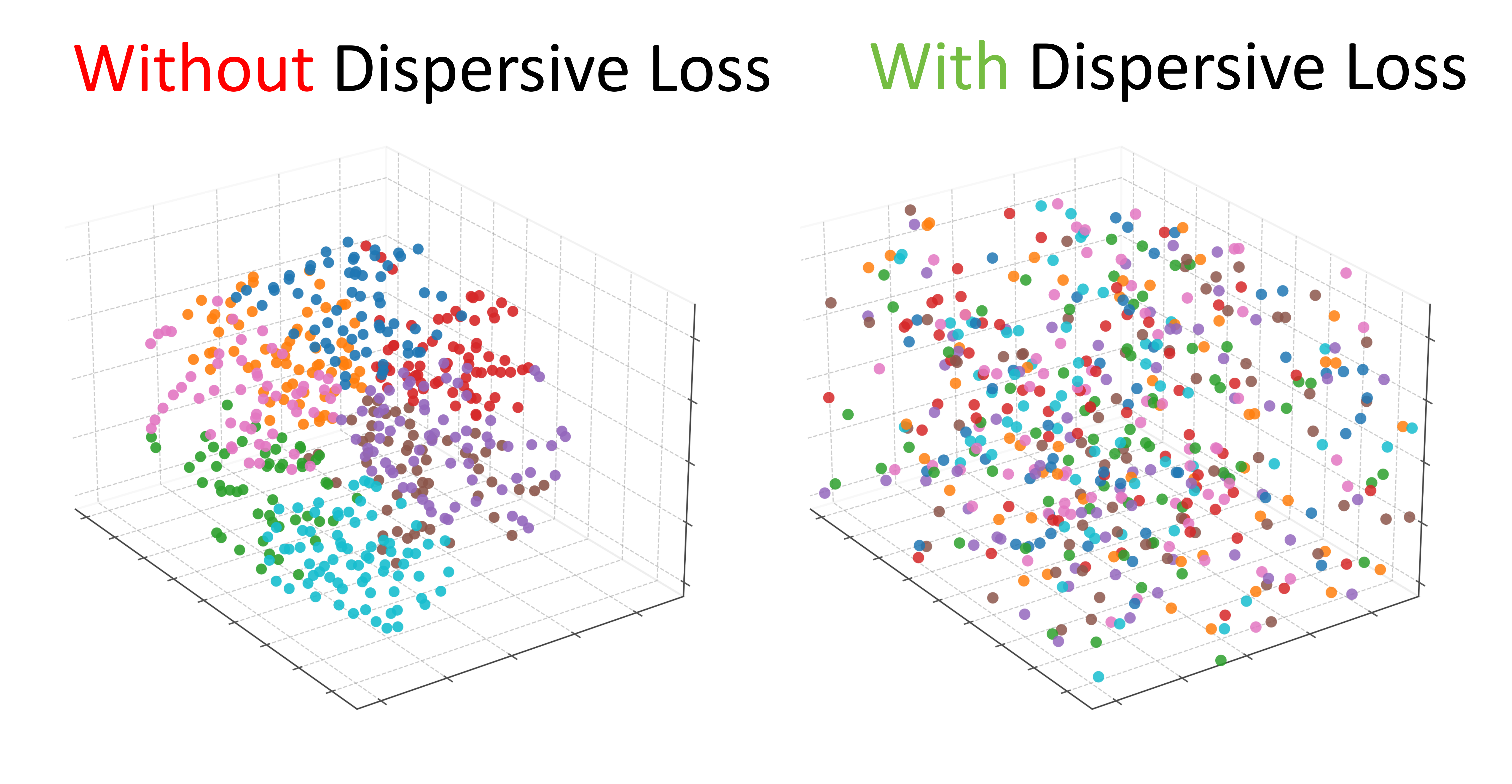}
    \vspace{-20pt}
    \caption{t-SNE visualization of latent representations with and without Dispersive Loss.}
    \label{fig:dis_tsne}
    \vspace{-20pt} 
\end{wrapfigure}
\textbf{Effect of Dispersive Loss.}
To qualitatively analyze the representation stabilization introduced by Dispersive Loss,
we visualize the latent feature distributions learned during Stage-1 pretraining.
As shown in Fig \ref{fig:dis_tsne}, without dispersive regularization, latent features exhibit noticeable collapse,
with samples clustered in a compact region.
In contrast, Dispersive Loss encourages a more uniformly distributed and
better separated latent space, indicating improved feature diversity and
training stability, consistent with the quantitative gains in Table~\ref{tab:ablation_pretrain}.

\textbf{Decoder designs in Stage-2.}
Table~\ref{tab:ablation_cbam} evaluates the decoder components on top of the pretrained baseline.
Both FDConv and DAF yield consistent improvements in overlap and boundary accuracy (FDConv: Dice $88.62\%\!\to\!91.65\%$, HD95 $17.24\!\to\!12.15$; DAF: Dice $88.62\%\!\to\!91.50\%$, HD95 $17.24\!\to\!12.34$), confirming the benefits of frequency--directional refinement and gated skip fusion.
In contrast, generic CBAM~\cite{woo2018cbam} does not improve Dice/mIoU and becomes detrimental when stacked with DAF, suggesting redundant reweighting that interferes with calibrated fusion.
Combining DAF and FDConv achieves the best overall performance (Dice $92.98\%$, HD95 $10.86$), demonstrating their complementarity for precise boundary delineation.

\subsection{Effect of Training Annotation Amount}
\label{sec:low_data}

\begin{wrapfigure}{r}{0.44\textwidth}
    \centering
    \vspace{-25pt} 
    \includegraphics[width=\linewidth]{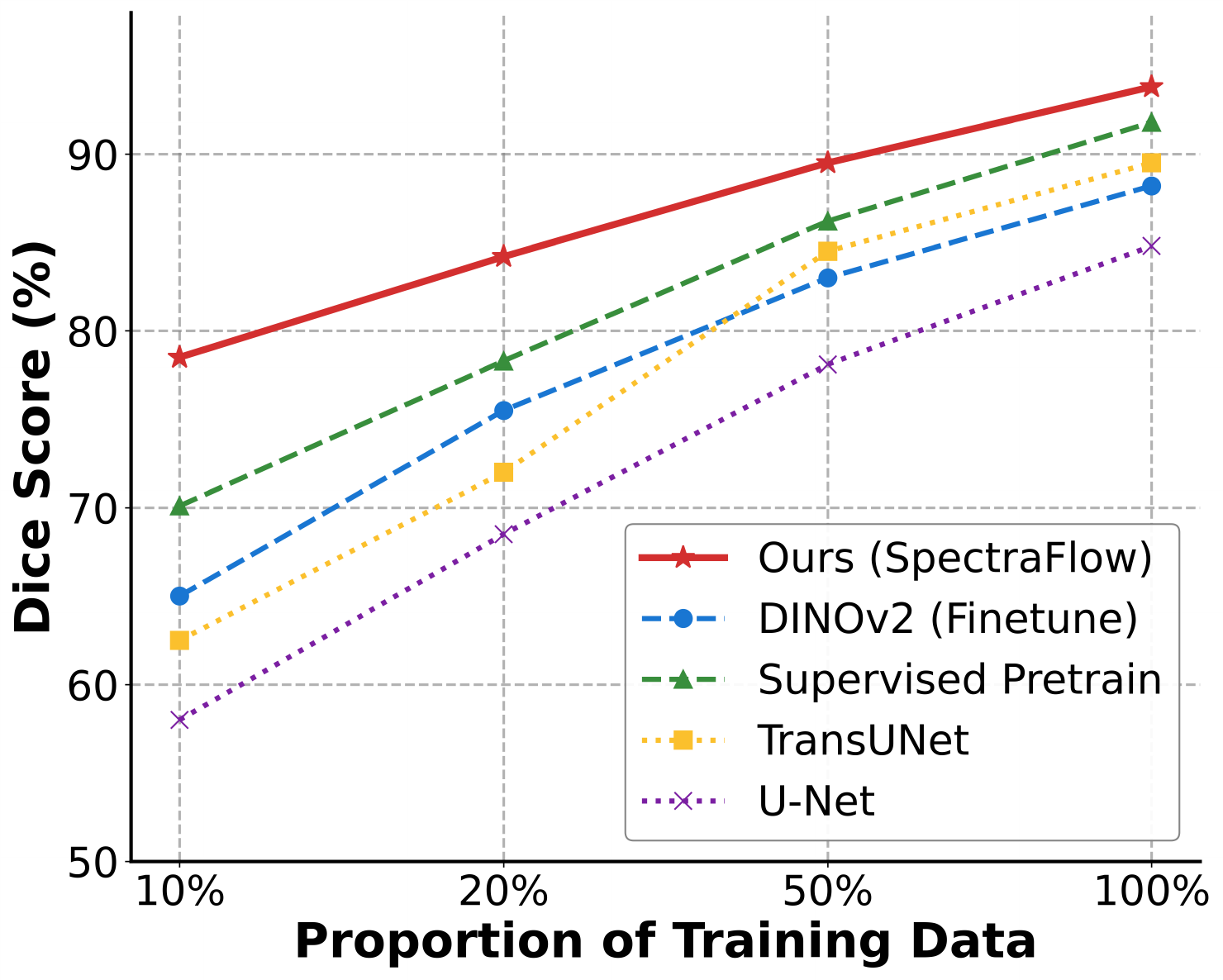}
    \caption{Effect of different proportions of training data on ISIC-2016.}
    \label{fig:low_data}
    \vspace{-15pt} 
\end{wrapfigure}

To evaluate data efficiency, we conduct experiments using different proportions of labeled training data on ISIC-2016.
Specifically, we randomly sample 10\%, 20\%, 50\%, and 100\% of the training set, while keeping the validation protocol unchanged.
For each setting, Stage-1 pretraining uses \emph{only} the masks from the same sampled subset as conditional structural inputs, ensuring that the amount of structural information available during pretraining is strictly aligned with the corresponding annotation ratio.

As shown in Fig.~\ref{fig:low_data}, our method consistently outperforms all baselines across all annotation ratios.
The advantage is particularly pronounced in low-data regimes.
With only 10\% labeled data, directly fine-tuning the official DINOv2 encoder leads to a substantial performance drop, whereas our geometry-guided latent pretraining maintains strong segmentation accuracy.

These results demonstrate that Stage-1 pretraining significantly improves representation generalization under limited supervision, enabling more reliable segmentation finetuning when annotated data is scarce.

\subsection{Robustness against Appearance Shifts}
\begin{figure}[t]
    \centering
    \includegraphics[width=0.65\textwidth]{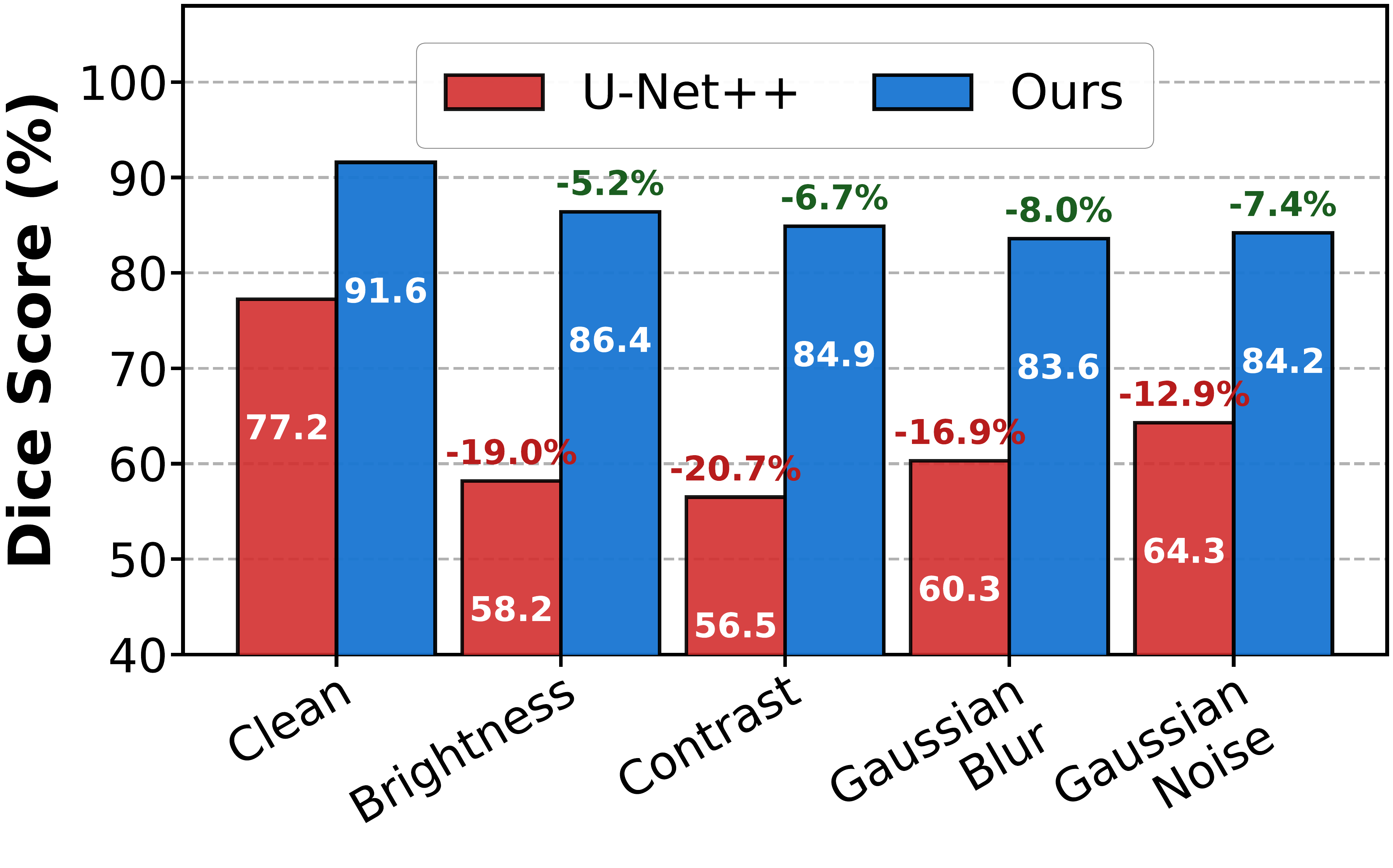}
    \caption{Robustness analysis against appearance corruptions on Kvasir-SEG.}
    \label{fig:shift}
\end{figure}
To further assess the robustness of our model under appearance perturbations, we subjected the test set to four types of corruptions: brightness, contrast, Gaussian blur, and Gaussian noise. 
As shown in Fig.~\ref{fig:shift}, the baseline U-Net++ exhibits notable performance degradation under these perturbations. 
In particular, under contrast variation, the baseline shows a Dice drop of 20.7\% (from 77.2\% to 56.5\%), suggesting a higher sensitivity to appearance changes. 
In contrast, our proposed SpectraFlow maintains a more stable performance, with a Dice decrease of only 6.7\% under the same condition. 
Across all corruption types, our method consistently outperforms the baseline, indicating that the structure-aware pretraining helps mitigate appearance sensitivity and promotes more structure-focused and appearance-robust representations.

\begin{table}[t!]
\centering
\caption{Organ-wise performance of different methods on the 3D Synapse dataset reported as DSC (\%).}
\label{tab:synapse_organ}

\resizebox{\textwidth}{!}{
\begin{tabular}{lcccccccc}
\toprule
\textbf{Organ} 
& \textbf{UNet} 
& \textbf{TransUNet} 
& \textbf{TransNorm} 
& \textbf{MT-UNet} 
& \textbf{SwinUNet} 
& \textbf{ConDSeg} 
& \textbf{WMREN} 
& \textbf{Ours} \\
& \footnotesize(MICCAI 2015) 
& \footnotesize(MICCAI 2022) 
& \footnotesize(IEEE Access 2022) 
& \footnotesize(IEEE ICIP 2020) 
& \footnotesize(ECCV 2022) 
& \footnotesize(AAAI 2025) 
& \footnotesize(IJCAI 2025) 
& \\
\midrule
Aorta        & 89.1 & 87.2 & 86.2 & 87.9 & 85.5 & 86.6 & 88.9 & 87.6 \\
Gallbladder  & 69.7 & 63.1 & 65.1 & 65.0 & 66.5 & 72.5 & 74.6 & 76.5 \\
Left kidney  & 77.8 & 81.9 & 82.2 & 81.5 & 83.3 & 85.1 & 88.5 & 81.6 \\
Right kidney & 68.6 & 77.0 & 78.6 & 77.3 & 79.6 & 78.6 & 84.2 & 84.3 \\
Liver        & 93.4 & 94.1 & 94.2 & 93.1 & 94.3 & 90.5 & 95.1 & 94.3 \\
Pancreas     & 54.0 & 55.9 & 55.3 & 59.5 & 56.6 & 61.2 & 69.6 & 72.6 \\
Spleen       & 86.7 & 85.1 & 89.5 & 87.8 & 90.7 & 87.3 & 91.1 & 91.2 \\
Stomach      & 75.6 & 75.6 & 76.0 & 76.8 & 76.6 & 79.8 & 83.1 & 80.9 \\
\midrule
\textbf{Mean DSC (\%) $\uparrow$} 
             & 76.9 & 77.5 & 78.4 & 78.6 & 79.1 & 80.2 & \underline{84.4} & \textbf{85.2} \\
\bottomrule
\end{tabular}
}
\end{table}

\subsection{Generalization Experiment on the 3D Synapse Dataset}
To assess the generalization capability and robustness of our method, we compare it with a set of representative CNN- and transformer-based segmentation approaches on the Synapse multi-organ dataset. Quantitative results measured by Dice Similarity Coefficient (DSC) are reported in Table~\ref{tab:synapse_organ}.

As shown in Table~\ref{tab:synapse_organ}, our method achieves the best overall performance with a mean DSC of 85.2\%, outperforming strong baselines such as UNet and TransUNet by +8.3 and +7.7 absolute points, respectively. Moreover, compared with recent methods including SwinUNet (79.1\%) and WMREN (84.4\%), our approach still yields a consistent gain on the mean score.

Notably, our method exhibits clear advantages on challenging organs with high shape variability and small volumes. In particular, it achieves 72.6\% DSC on the Pancreas and 76.5\% DSC on the Gallbladder, surpassing the second-best method by 3.0 and 1.9 points, respectively. These results suggest that the proposed design better preserves fine-grained boundary details and improves segmentation reliability for small and anatomically complex structures, contributing to the overall performance gain.

\begin{figure}[t]
  \centering
  \includegraphics[width=0.95\textwidth, trim=0 10 0 5, clip]{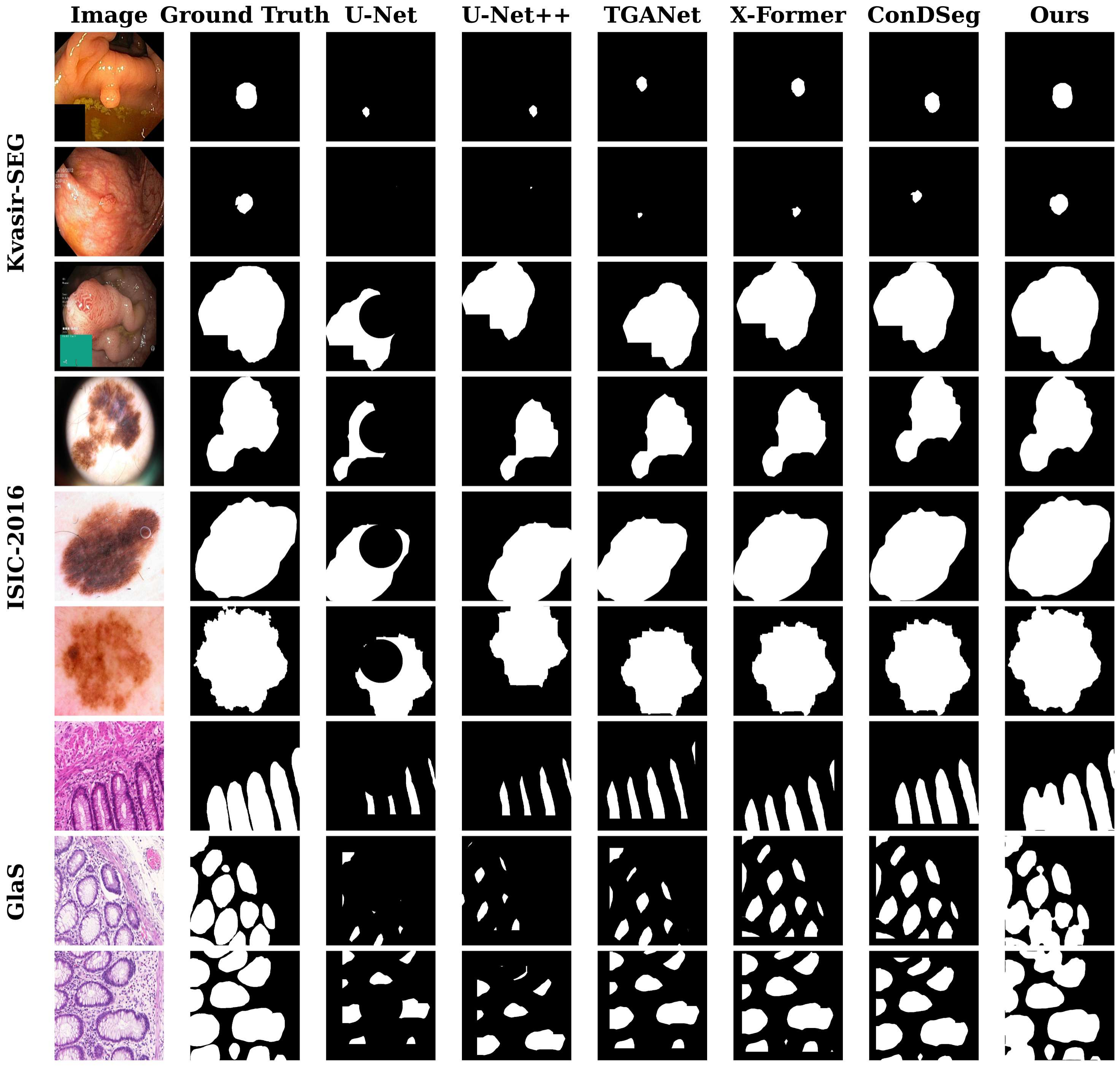}
  \caption{Qualitative comparison of segmentation results on Kvasir-SEG, ISIC-2016, and GlaS.}
  \label{fig:qualitative}
\end{figure}

\subsection{Qualitative Comparison}

Figure~\ref{fig:qualitative} presents qualitative segmentation results on Kvasir-SEG, ISIC-2016, and GlaS.
Conventional CNN-based methods (e.g., U-Net and U-Net++) tend to miss small targets or produce fragmented predictions, while transformer-based models often yield over-smoothed boundaries.
In contrast, our method consistently preserves fine structures and produces more coherent object shapes across different modalities, particularly in challenging cases with small lesions or complex boundaries.
These visual results are consistent with the quantitative improvements reported in Table~\ref{tab:compare1}, and further demonstrate the robustness and generalization capability of the proposed framework.

\label{sec:qualitative}
\begin{wrapfigure}{r}{0.35\textwidth}
    \centering
    \vspace{-20pt}
    \includegraphics[width=\linewidth]{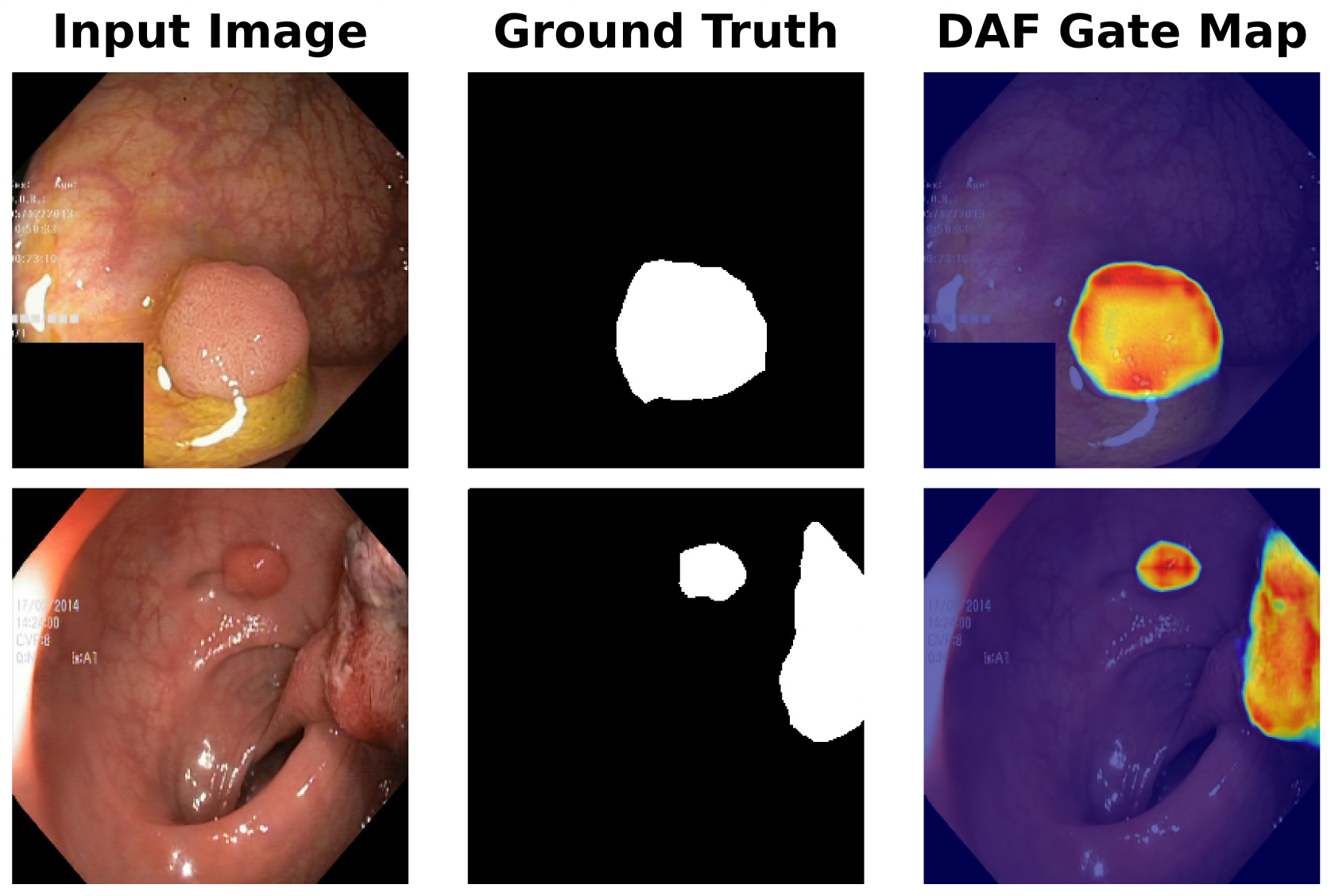}
    \vspace{-20pt}
    \caption{
    Qualitative visualization of DAF gating behavior.
    }
    \label{fig:daf_gate}
\end{wrapfigure}

\textbf{Effect of DAF Gated Fusion.}
To analyze the contribution of Direct Attentional Fusion (DAF) in Stage-2, we visualize the learned gate map $\mathbf{M}$ produced by the sigmoid activation. As shown in Fig.~\ref{fig:daf_gate}, DAF assigns higher responses to boundary-relevant and lesion regions, while suppressing background and irrelevant textures. This content-adaptive gating mechanism helps bridge the semantic gap between encoder and decoder features, leading to more accurate boundary delineation.

\section{Conclusion}
\label{sec:conclusion}
This paper presents a two-stage framework that shifts the focus from texture-biased patterns to geometric structures in medical image segmentation. Stage-1 introduces MeanFlow-based latent transport pretraining to establish a geometry-consistent representation manifold. We incorporate a lightweight Dispersive Loss to regularize the latent space and effectively mitigate representation collapse. In Stage-2, the decoder leverages Direct Attentional Fusion and Frequency-Directional Dynamic Convolution to achieve precise boundary refinement. Extensive evaluations across multiple benchmarks demonstrate our superior robustness and data efficiency, particularly in low-annotation regimes.

%
%
\bibliographystyle{splncs04}
\bibliography{main}

\appendix
\section{Extra Experiments}
\begin{figure}[t!] 
    \centering
    \includegraphics[
        width=\linewidth,
        trim=0cm 0cm 0cm 0cm,
        clip
    ]{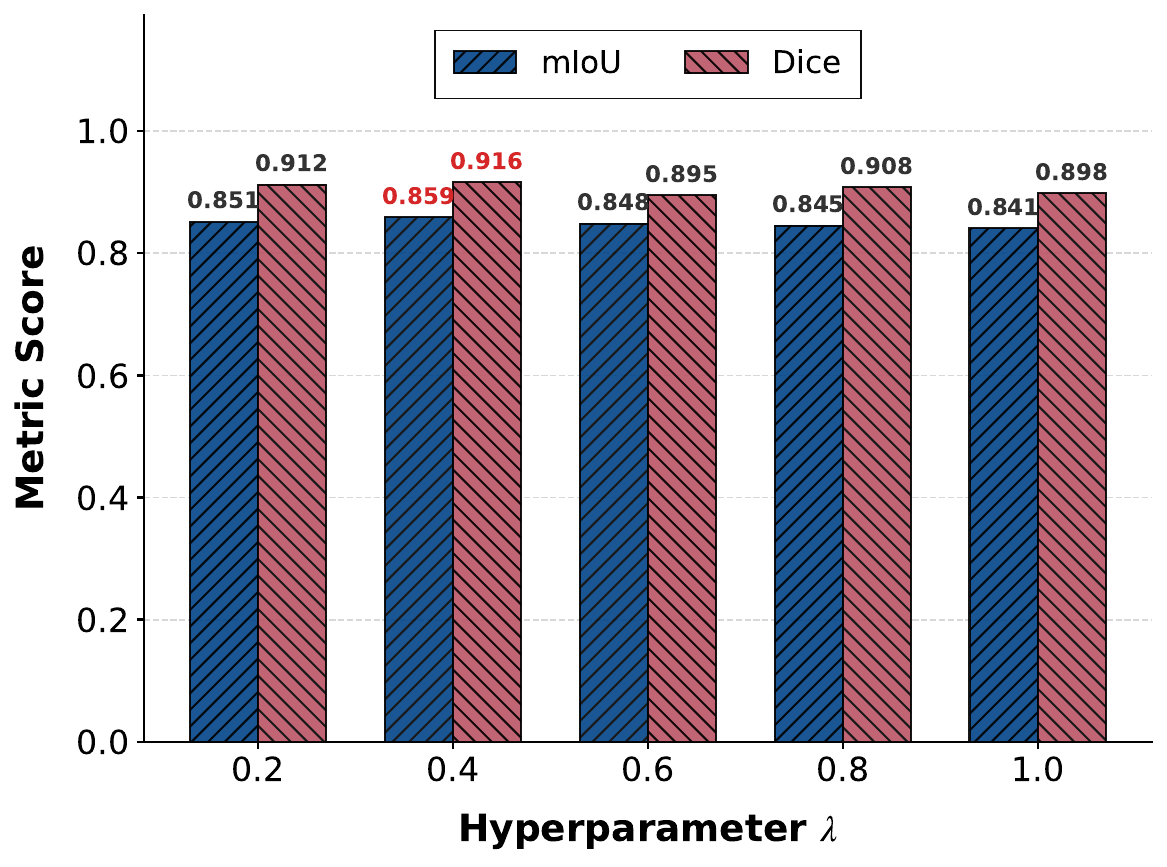}
    \caption{Sensitivity Analysis of the Dispersive Loss Weight
    }
    \label{fig:disweight}
\end{figure}
\subsection{Sensitivity Analysis of the Dispersive Loss Weight}
To investigate the influence of the Dispersive Loss weight, we conduct a sensitivity analysis on Kvasir-SEG dataset by varying the hyperparameter $\lambda$ from 0.2 to 1.0. The quantitative results in terms of mIoU and Dice score are visualized in Fig.~\ref{fig:disweight}.

As shown in the figure, the model performance initially improves and peaks at $\lambda=0.4$, yielding an mIoU of 0.859 and a Dice score of 0.916. The performance remains relatively stable across different values, indicating that our framework is robust to the choice of $\lambda$. However, when $\lambda$ increases further, a slight performance drop is observed. This suggests that while dispersive regularization is beneficial, an excessively strong constraint may overly disperse the latent space and limit feature adaptability.

\begin{table}[t]
\centering
\caption{Ablation study on Stage-2 encoder fine-tuning strategies on ISIC-2016. We compare the performance of keeping the encoder frozen, fully fine-tuning it, and fine-tuning only the last block (Ours).}
\label{tab:finetune_strategy}
\setlength{\tabcolsep}{10pt}
\begin{tabular}{lcc}
\toprule
\textbf{Fine-tuning Strategy} & \textbf{DSC (\%)}~$\uparrow$ & \textbf{HD95}~$\downarrow$ \\
\midrule
Frozen Encoder & 91.28 & 12.52 \\
Full Fine-tuning & 89.72 & 16.68 \\
Last Block Fine-tuning (Ours) & \textbf{92.98} & \textbf{10.86} \\
\bottomrule
\end{tabular}
\end{table}
\subsection{Ablation Study on Stage-2 Encoder Fine-tuning Strategies}

To determine an effective adaptation strategy for the pre-trained encoder in Stage-2, we compare three fine-tuning protocols on the ISIC-2016 dataset: (1) keeping the encoder frozen, (2) fully fine-tuning the entire encoder, and (3) fine-tuning only the last encoder block. The quantitative results are summarized in Table~\ref{tab:finetune_strategy}.

As shown in the table, fully fine-tuning the encoder leads to a noticeable performance drop, with DSC decreasing from 91.28\% (Frozen) to 89.72\%. This suggests that aggressive fine-tuning on limited medical data may disrupt the structural representations learned during Stage-1 pre-training. In contrast, keeping the encoder frozen preserves these representations but limits the model's ability to adapt high-level features to the target segmentation task.

Fine-tuning only the last encoder block achieves the best performance, reaching a DSC of 92.98\% and an HD95 of 10.86. This result indicates that partial fine-tuning provides a favorable trade-off: it retains low-level structural information while allowing deeper layers to better align with task-specific semantic cues.

\section{From InfoNCE to Dispersive Loss}
The InfoNCE objective encourages high similarity for a designated positive pair $(z_i, z_i^{+})$ while reducing similarity to all other samples. A common form is
\begin{equation}
\mathcal{L}_{\text{InfoNCE}}
= - \log \left(
\frac{\exp\!\left(-\frac{D(z_i,z_i^{+})}{\tau}\right)}
{\exp\!\left(-\frac{D(z_i,z_i^{+})}{\tau}\right)+\sum_{j\neq i}\exp\!\left(-\frac{D(z_i,z_j)}{\tau}\right)}
\right),
\label{eq:info_nce}
\tag{12}
\end{equation}
or equivalently
\begin{equation}
\mathcal{L}_{\text{InfoNCE}}
= - \log \left(
\frac{\exp\!\left(-\frac{D(z_i,z_i^{+})}{\tau}\right)}
{\sum_{j=1}^{N}\exp\!\left(-\frac{D(z_i,z_j)}{\tau}\right)}
\right).
\tag{13}
\end{equation}
Here $D(\cdot,\cdot)$ denotes a dissimilarity measure and $\tau>0$ is the temperature; the denominator pools both the positive and all negatives.

Using elementary log identities, InfoNCE can be rearranged into two parts:
\begin{equation}
\label{eq:info_nce_decomposed}
\begin{aligned}
\mathcal{L}_{\text{InfoNCE}}
&= \frac{D(z_i, z_i^{+})}{\tau}
  \;+\;
  \log\!\left(
      \exp\!\left(-\frac{D(z_i,z_i^{+})}{\tau}\right)
      \right)
\\[6pt]
&\quad
+\,
\sum_{j\neq i}\exp\!\left(-\frac{D(z_i,z_j)}{\tau}\right).
\end{aligned}
\tag{14}
\end{equation}
where the first term rewards the positive pair, and the second is a log-sum-exp normalization over the batch.

Motivated by the latter term, we retain only this normalization to promote dispersion without relying on explicit positives:
\begin{equation}
\mathcal{L}_{\text{Disp}}
= \log \sum_{j=1}^{N} \exp\!\left(-\frac{\mathcal{D}(z_i,z_j)}{\tau}\right).
\label{eq:disp-base}
\tag{15}
\end{equation}
For a usable batch objective, we average over anchors:
\begin{equation}
\mathcal{L}_{\text{Disp}}
= \mathbb{E}_{i}\!\left[
\log \sum_{j=1}^{B}
\exp\!\left(-\frac{\mathcal{D}(z_i,z_j)}{\tau}\right)
\right],
\label{eq:disp-batch}
\tag{16}
\end{equation}
which drives all representations in the batch to repel one another and mitigates representational collapse.

\textbf{Dispersive Loss Variants}\quad
\textbf{InfoNCE L2 Variant.} With squared Euclidean distance $\mathcal{D}(h_i,h_j)=\lVert h_i-h_j\rVert_2^{2}$ and excluding $i=j$,
\begin{equation}
\mathcal{L}_{\text{disp}}^{\text{L2}}
=
\log\!\left(
\frac{1}{B(B-1)}
\sum_{i=1}^{B}\sum_{\substack{j=1\\ j\ne i}}^{B}
\exp\!\left(-\frac{\lVert h_i-h_j\rVert_2^{2}}{\tau}\right)
\right).
\label{eq:disp-l2}
\tag{17}
\end{equation}

\textbf{InfoNCE Cosine Variant.} Using cosine dissimilarity
$\mathcal{D}(h_i,h_j)=1-\frac{h_i^{\top}h_j}{\lVert h_i\rVert_2\,\lVert h_j\rVert_2}$,
\begin{equation}
\mathcal{L}_{\text{disp}}^{\text{cos}}
=
\log\!\left(
\frac{1}{B(B-1)}
\sum_{i=1}^{B}\sum_{\substack{j=1\\ j\ne i}}^{B}
\exp\!\left(
-\frac{\,1-\frac{h_i^{\top}h_j}{\lVert h_i\rVert_2\,\lVert h_j\rVert_2}\,}{\tau}
\right)
\right).
\label{eq:disp-cos}
\tag{18}
\end{equation}

\textbf{Hinge Loss Variant.} Enforcing a margin $\epsilon>0$ between representations:
\begin{equation}
\mathcal{L}_{\text{disp}}^{\text{hinge}}
=
\frac{1}{B(B-1)}
\sum_{i=1}^{B}\sum_{\substack{j=1\\ j\ne i}}^{B}
\max\!\Bigl(0,\;\epsilon-\lVert h_i-h_j\rVert_2^{2}\Bigr)^{2}.
\label{eq:disp-hinge}
\tag{19}
\end{equation}

\textbf{Covariance Off-Diagonal Penalty.} Encouraging decorrelation across feature dimensions:
\begin{equation}
\label{eq:disp-cov}
\begin{aligned}
\mathcal{L}_{\text{disp}}^{\text{cov}}
&=
\left\|C-\operatorname{diag}(C)\right\|_{F}^{2},
\qquad
C=\frac{1}{B-1}\tilde H^{\top}\tilde H,
\\[6pt]
&\quad
\tilde H = H - \mathbf{1}\,\bar h^{\top},
\qquad
\bar h = \tfrac{1}{B}\sum_{i=1}^{B}h_i .
\end{aligned}
\tag{20}
\end{equation}

\paragraph{Mathematical Properties of Dispersive Loss — Gradient Analysis.}
The gradient with respect to an anchor $h_i$ takes a weighted repulsive form:
\begin{equation}
\label{eq:disp-grad}
\begin{aligned}
\frac{\partial \mathcal{L}_{\text{Disp}}}{\partial h_i}
&=
\frac{1}{\tau}\sum_{j=1}^{B}
w_{ij}\,
\frac{\partial \mathcal{D}(h_i,h_j)}{\partial h_i},
\\[6pt]
w_{ij}
&=
\frac{\exp\!\left(-\frac{\mathcal{D}(h_i,h_j)}{\tau}\right)}
{\sum_{\ell=1}^{B}\exp\!\left(-\frac{\mathcal{D}(h_i,h_\ell)}{\tau}\right)} .
\end{aligned}
\tag{21}
\end{equation}
For $\mathcal{D}(h_i,h_j)=\lVert h_i-h_j\rVert_2^{2}$ this reduces to
\begin{equation}
\frac{\partial \mathcal{L}_{\text{Disp}}}{\partial h_i}
=
\frac{2}{\tau}\sum_{j=1}^{B} w_{ij}\,(h_i-h_j),
\label{eq:disp-grad-l2}
\tag{22}
\end{equation}
so closer neighbors (with larger $w_{ij}$) exert stronger repulsion, naturally spreading batch representations.
\end{document}